\documentclass[sigconf,screen]{acmart}
\AtBeginDocument{%
  \providecommand\BibTeX{{%
    \normalfont B\kern-0.5em{\scshape i\kern-0.25em b}\kern-0.8em\TeX}}}

\setcopyright{acmcopyright}
\copyrightyear{2021}
\acmYear{2021}

\acmConference[SimAUD 2021]{SimAUD 2021: Symposium on Simulation for Architecture and Urban Design}{April 15--17, 2021}{Online}
\acmBooktitle{SimAUD 2021: ASymposium on Simulation for Architecture and Urban Design, April 15--17, 2021, Online}
\acmPrice{15.00}
\acmISBN{978-1-4503-XXXX-X/21/04}

\acmSubmissionID{38}

\begin{document}

\title{Generative Adversarial Networks for Urban Block Design}

\author{Stanislava Fedorova}
\authornotemark[1]
\email{stanislava.fedorova@mail.polimi.it}

\affiliation{%
  \institution{Politecnico di Milano}
  \streetaddress{Piazza Leonardo da Vinci, 32}
  \city{Milan}
  \state{MI}
  \country{Italy}
  \postcode{20133}
}

\renewcommand{\shortauthors}{Fedorova, et al.}

\begin{abstract}
Development and diffusion of machine learning and big data tools provide a new tool for architects and urban planners that could be used as analytical or design instruments. The topic investigated in this paper is the application of Generative Adversarial Networks to the design of an urban block. The research presents a flexible model able to adapt to the morphological characteristics of a city. This method does not define explicitly any of the parameters of an urban block typical for a city, the algorithm learns them from the existing urban context. This approach has been applied to the cities with different morphology: Milan, Amsterdam, Tallinn, Turin, and Bengaluru in order to see the performance of the model and the possibility of style translation between different cities. The data are gathered from Openstreetmap and Open Data portals of the cities. This research presents the results of the experiments and their quantitative and qualitative evaluation.
\end{abstract}




\maketitle

\section{Introduction}
Urban design solutions determine many important aspects of urban life: economical, ecological, social aspects, transportation, pedestrian flows, consequently, multiple parameters are involved. The necessity to consider many different factors interacting with each other made this problem a challenge in the field of generative design. The result of an urban project has a long-lasting effect on various spheres of life, which makes finding an optimal solution even more important. The issue arises when it is necessary to define a set of parameters determining an optimal solution for a given area. Some of them are dictated by the local regulations, some by common knowledge, some by the sense of aesthetics; sometimes it is difficult to formulate them using words while algorithms would require formulae or numeric representation.

One of the problems involving a manifold of parameters is a design of an urban block. This problem takes into consideration the shapes of the buildings, their proportions, relationships with the adjacent buildings, the composition of the open space, relationships of buildings with the open space, with the urban block itself, relationships with the buildings of the adjacent blocks. Block design involves the knowledge of the typologies of the blocks in a city, surroundings (streets, squares, railways nearby), etc. Moreover, it should smoothly integrate into the existing urban context, which means taking into consideration the disposition and typology of the adjacent blocks. All of the parameters mentioned above would be difficult to define for each block a design should be produced to.
 
With the recent introduction of Generative Adversarial Networks (GANs)\cite{goodfellow2014generative}, it became possible to use the neural networks' ability to autonomously learn visual features in order to generate new content. The main advantage of this technique applied to generative design is the elimination of the necessity to determine any parameters of the desired design solution and learning the visual features determining the existing city structure and consequently generating blocks imitating this structure.
 
\section{Related Work}
As the concept of GANs has been introduced quite recently\cite{goodfellow2014generative} there has not been much research on the topic concerning architecture and urban design, although it is well-known in the field of computer science.

Prior work in the generative urban design field is mostly based on a set of defined rules and constraints according to which several design options are generated\cite{rule_based_design}. Nagy, Villaggi, and Benjamin\cite{inproceedings} explore optimization of the problem complexity with the set of two principal goals and the use of genetic algorithms. However, their approach, too, needs a set of constraints and requirements; their model outputs a design space out of which, after optimization, only a few designs are chosen as final.

The use of GANs has been explored in interior design as well as in visualization\cite{nvidia} and simulation. Stanislas Chaillou\cite{archigan} explored the functionality of the method applied to the design of the apartments generating their shape, internal structure, functional zoning, and furniture placement. The same approach at the same architectural scale\cite{daylightgan} was used for the simulation of natural illumination in an apartment provided its plan. Another work published in 2020 used a GAN with a different architecture with respect to ours for landuse planning\cite{landuseai} with the use of predefined parameters that a planner considers important for his solution and the zone design. However, this approach produces only a large-scale distribution of the functional zoning in the city and is not applicable to the urban design.
\newline The present work tackles a different scale problem: urban block level design. Instead of using a set of rules exploited by the generative urban design models previously, the GAN learns the set of parameters present in the city block design from the visual features of the images. A work that uses GANs in architecture on a scale that is similar to ours is the project This Map Does Not Exist\cite{thismapdoesnotexist}. However, this work uses a different architecture that does not allow the production of the designs based on the existing input: the maps are not attached to any existing point in the real world and the existing conditions of an urban environment are not taken into account when producing an image. One more work\cite{ntg} uses generative networks for the generation of an urban street layout. This approach is the closest to our work both in terms of urban scale and the possibility of conditional generation. The problem is explored from a different point of view: the authors use graph generation to produce a road design while our research tackles the generation of the urban blocks.
\newline Importantly, our research combines both the challenges faced in the previously mentioned works using GANs: production of a design solution based on the chosen parameters of the surrounding environment and the integrity of the solution within its urban block.

\section{Data sampling and acquisition}
Firstly, the data of the city of Milan was used to make a dataset. This city has been chosen due to its characteristic morphology and relatively consistent urban fabric inside the external ring road and within a 1 km distance from it.

The dataset consists of images that are diagrammatic representations of the city zones. It has been decided to start with the basic parameters determining the composition of a building block: building footprints, building heights' categories, road structure, and railways. The heights of the buildings have been divided into 3 ranges: low-rise structures, medium-rise structures, and high-rise structures depending on the distribution of the heights in the city. Heights in the diagrammatic representation are defined as three gray values. In case of height not having been mentioned for a building, a mean value for the city was used to denote its height.

The dataset has been generated with the use of QGIS software \cite{QGIS_software} with the condition to have more than 12 percent of a block of interest occupied. Urban blocks containing exclusively industrial buildings have been excluded from the datasets as their presence introduces additional variance. Since the information about the landuse and the building type is not provided to the algorithm the experiment is concentrated on the buildings with residential and mixed functions. Moreover, the proportions of the industrial buildings are similar across different geographical regions and do not introduce any characteristics typical to a specific city.

The images were produced in scale 1:3000 in RGB colorspace; these images were scaled down from 2100x2100 to 256x256 pixels to be fed into the model.

\begin{figure}[!h]
    \centering
    \includegraphics[width=1.0\columnwidth]{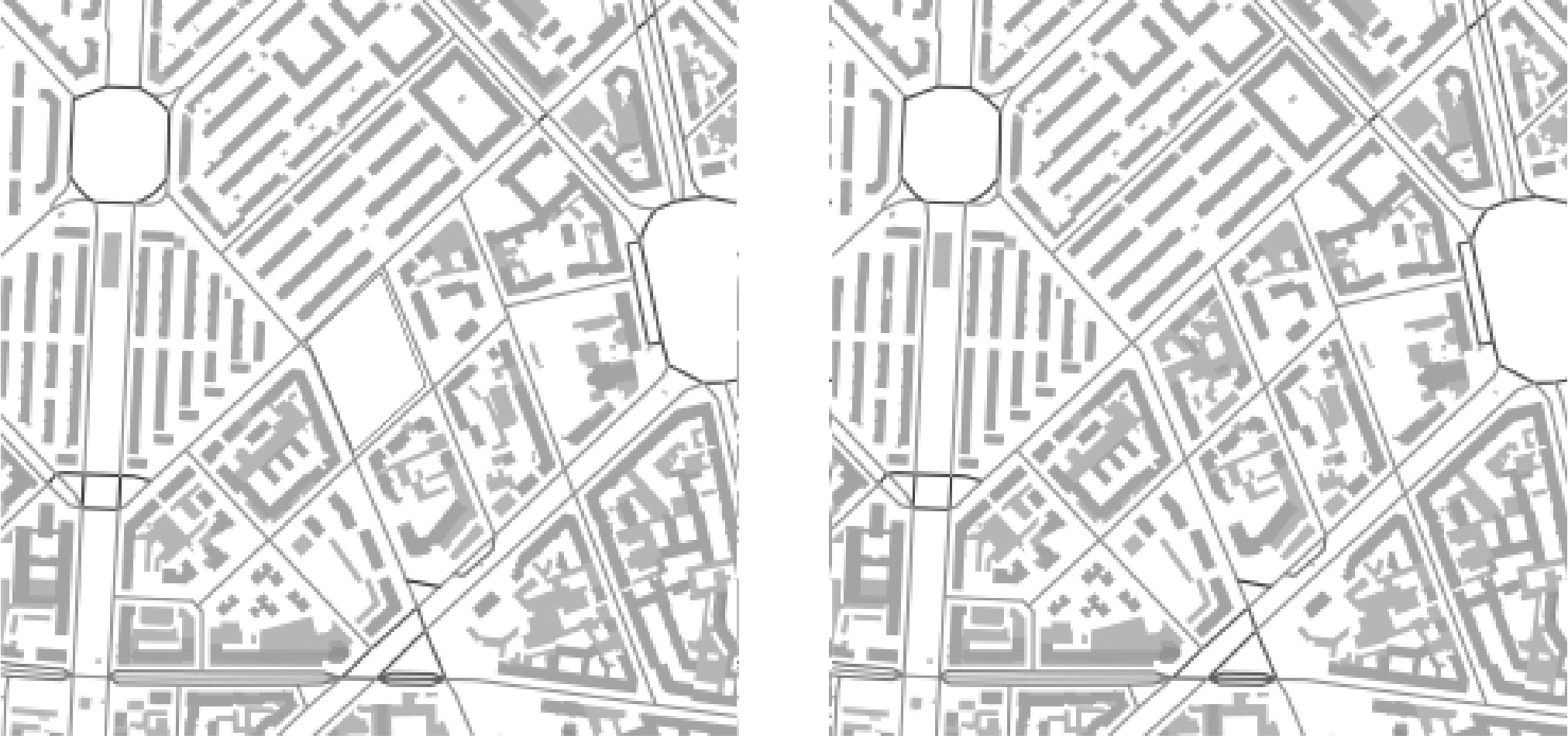}
    \caption{Sample from Milan Dataset of urban blocks: on the left an image from set A, on the right a corresponding image from set B. The white block in the center of the left image is the block to be designed.}
    \label{fig:figure1}
    \Description{Two images showing a diagrammatic representation of the top view on a part of the city of Milan (Piazzale Segesta). The image on the left is the input condition, the image on the right is the ground truth.}
\end{figure}

The dataset consists of two sets: set B includes the diagrammatic representations of the city and set A holds the same images with an empty polygon in the center of the image, destined to be filled with the buildings as in Figure~\ref{fig:figure1}. Code for dataset generation is available in the repository\cite{github}.

The data was acquired from open sources: Open Data Portal of Comune di Milano \cite{Milano_geoportale} available under CC License \cite{cc_license} and Openstreetmap.org \cite{OpenStreetMap} available under the Open Database Licence. The absent parameters have been considered as a mean value for the city. No modifications have been made to the datasets provided. 

\begin{figure*}[th!]
    \centering
    \includegraphics[width=1.0\textwidth]{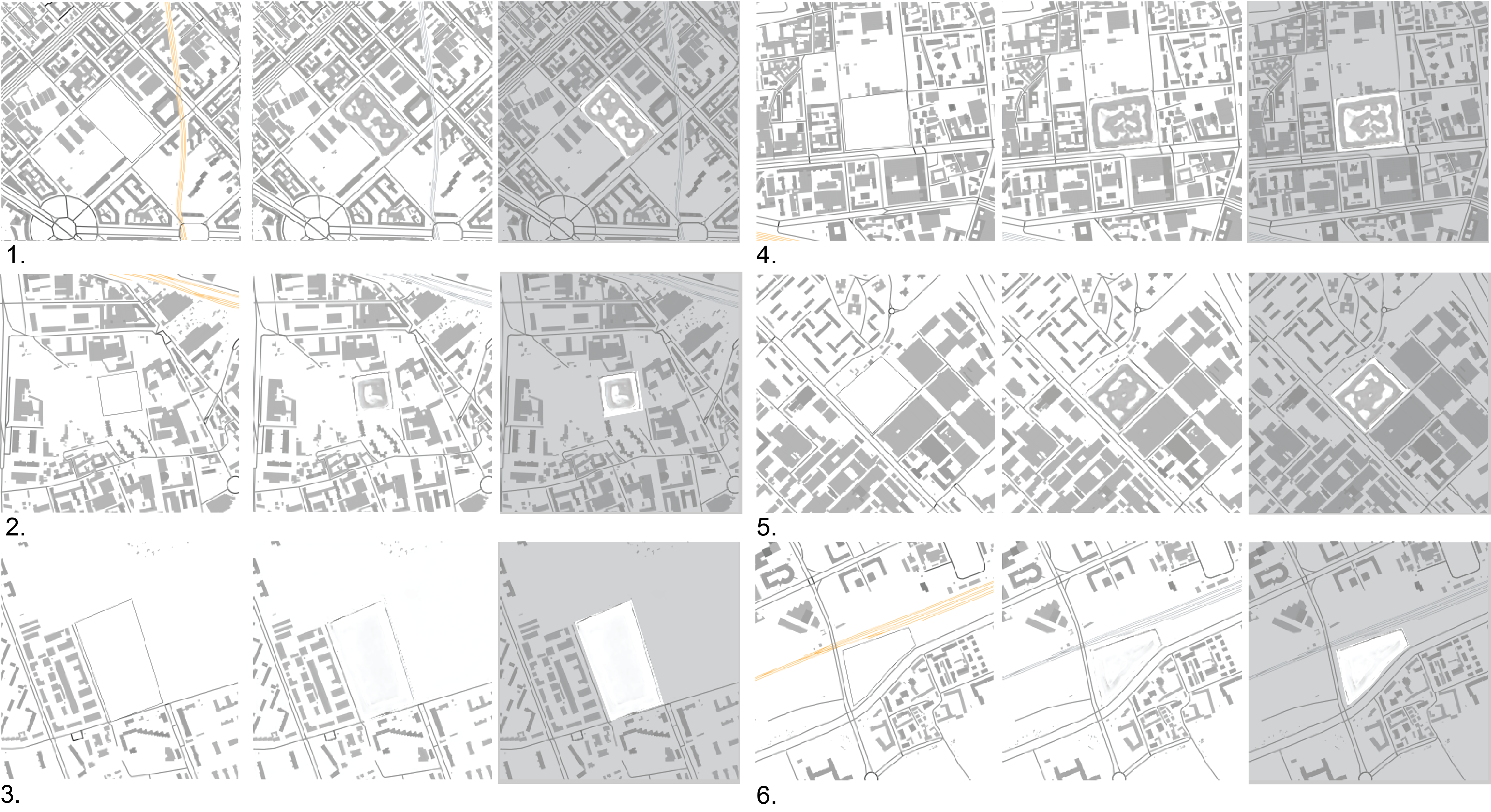}
    \centering
    \caption{The results were obtained on the test set of Milan images with the model trained on Milan urban blocks. The images are presented in sets of 3: input, output, output with the zone of interest highlighted (for the clarity of reading the results). The zones have been taken from the Riformare Milano project\cite{riformaremilano}. 1 - Dergano zone; 2 - Oriani zone; 3 - Caserma Perucchetti; 4 - Dergano zone; 5 - Mecenate zone; 6 - San Cristoforo zone}
    \label{fig:milan_test}
    \Description{A 3x6 matrix showing 6 samples from the Milan test set results: the first column is the input image, the second is the output from the model and the third is the ground truth. The images are diagrammatic representations of the city plan.}
    \centering
\end{figure*}

\section{Architecture}
The architecture used for this research is a GAN for Image-to-Image translation \cite{isola2018imagetoimage}. The algorithm consists of two models: Generator Network based on Unet architecture and Discriminator Network based on PatchGAN Convolutional Network. The principle is in mutual learning: the generator generates an output image based on the condition (input image). The discriminator model receives generated images and target images and learns to distinguish between them, classifying each of them into a 'real' or a 'fake' class. In the case of PatchGAN, the model learns to classify patches of a given image as real or fake instead of giving a result for the whole image sample, so it outputs the result over which areas of an image should be improved by the generator.

Three different implementations of the algorithm were used, which are: Tensorflow-based \cite{pixtf}, PyTorch-based without Unet \cite{pixtorch}, Pytorch-based with Unet \cite{github} \cite{ganspecialization}. The use of adaptive learning rate tends to give better results. An algorithm compares the mean loss from 250 iterations with the previous accumulated losses. In case the latest loss is larger than the largest of the previous three losses, the learning rate reduces. This technique results in more distinct shapes of the building contours and more efficient training in terms of time. The algorithm has been trained using Google Colab environment with a single NVIDIA Tesla K80 GPU.

\section{Generation of Milan Urban Style}
The specifics of Milan morphology and consistency in the urban fabric of the center of the city make it a good location for the initial experiments. The relative homogeneity of the urban morphology reduces the variance in the data. The model has been trained for 20 epochs with a dataset of 1200 images.

After the training, the model has been tested on a new set of images that were not included in the training set. The areas chosen for the test have been previously defined as 'problematic'\cite{riformaremilano} and requiring a transformation. However, the main point is not a proposal of a solution to these areas, but a demonstration of the possibilities of GAN-based generative urban design, even in the complex urban contexts that are different from the residential ones. The results in Figure~\ref{fig:milan_test} show designs of the urban blocks coherent with the traditional Milanese urban blocks\cite{Biraghi:2018:ISBN:978-5-7638-4127-5} that tend to have a closed structure and an internal courtyard system in their majority. The model successfully recognizes the zone of interest to construct the block within and manages to adapt the learned visual characteristics of the city into a new context. Samples 1, 2, 4 are the samples with the irregular urban fabric and sample 5 is an example of an urban context with a high presence of industrial buildings differing in proportions from the forms in the training set.

At the same time model fails to produce a clear solution in a few samples (3, 6 in Figure~\ref{fig:milan_test}). This behavior occurs in the samples having a low urban density of the surroundings or their absence. Another reason is connected to the absence of the surroundings: their small proportion in the image to the zone of interest (e.g. when the zone occupies more than 30 percent of an image). This factor is an indication of the fact that the model produces solutions based on the visible surroundings; at the same time, it introduces a limitation of its use as well as a constraint for the input images: the model is not capable of producing design without any surroundings. This behavior is consistent on different urban scales and is related mostly to the proportion of the block to the image.
\begin{figure}[h!]
    \centering
    \includegraphics[width=0.95\columnwidth]{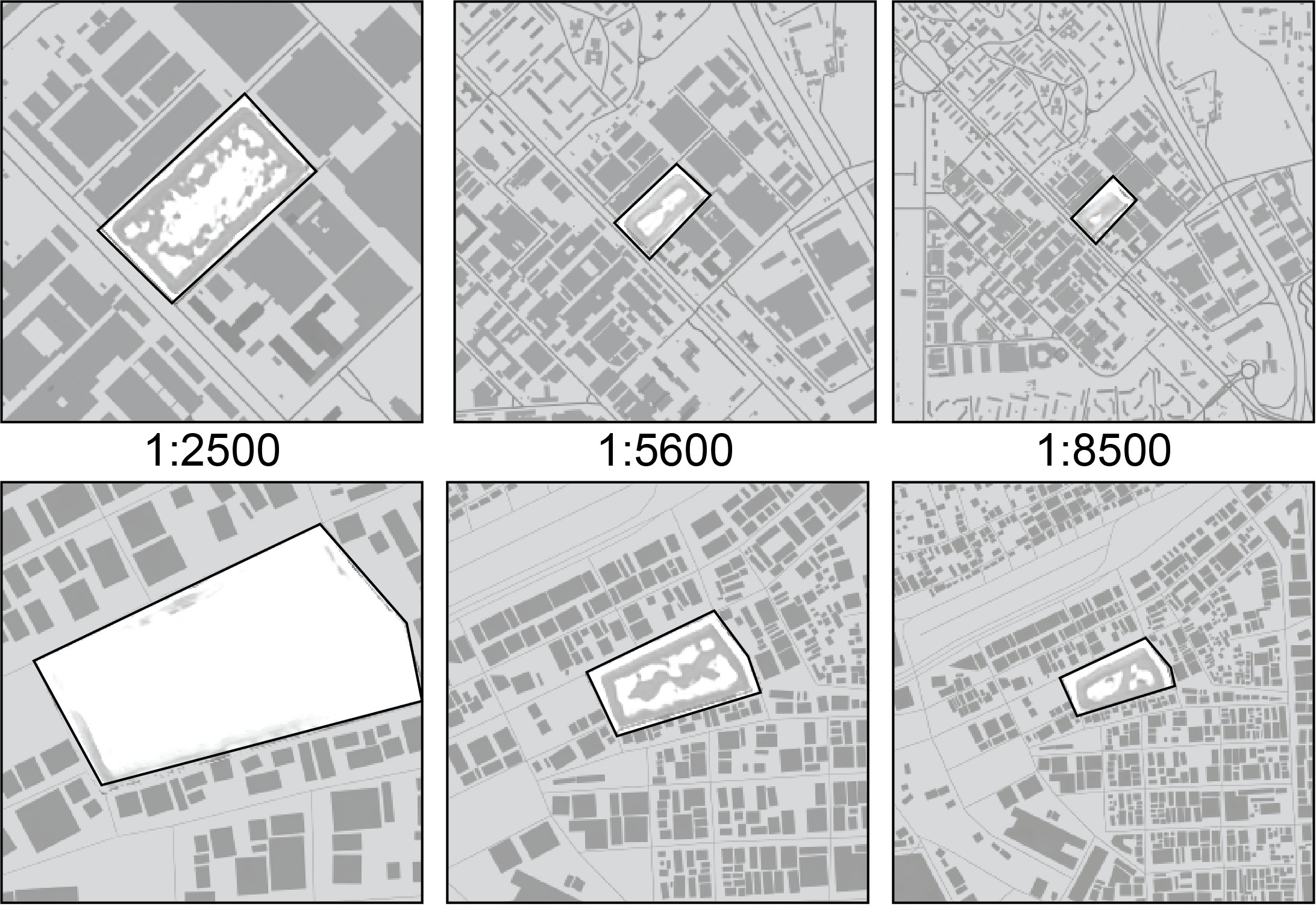}
    \caption{Illustration of the model's performance on the urban contexts in different scales. The images are on scales 1:2500, 1:5600, and 1:8500 respectively; the upper row represents the city of Milan, Mecenate zone; the second row is the city of Bengaluru.}
    \label{fig:scales}
    \Description{A 2x3 matrix with a diagrammatic representation of Milan and Bengaluru top views in different scales (1:2500, 1:5600, 1:8500) and with a different area of the zone of interest within one scale with the results from the generative models.}
    \centering
\end{figure}
\begin{figure}[h!]
    \centering
    \includegraphics[width=0.95\columnwidth]{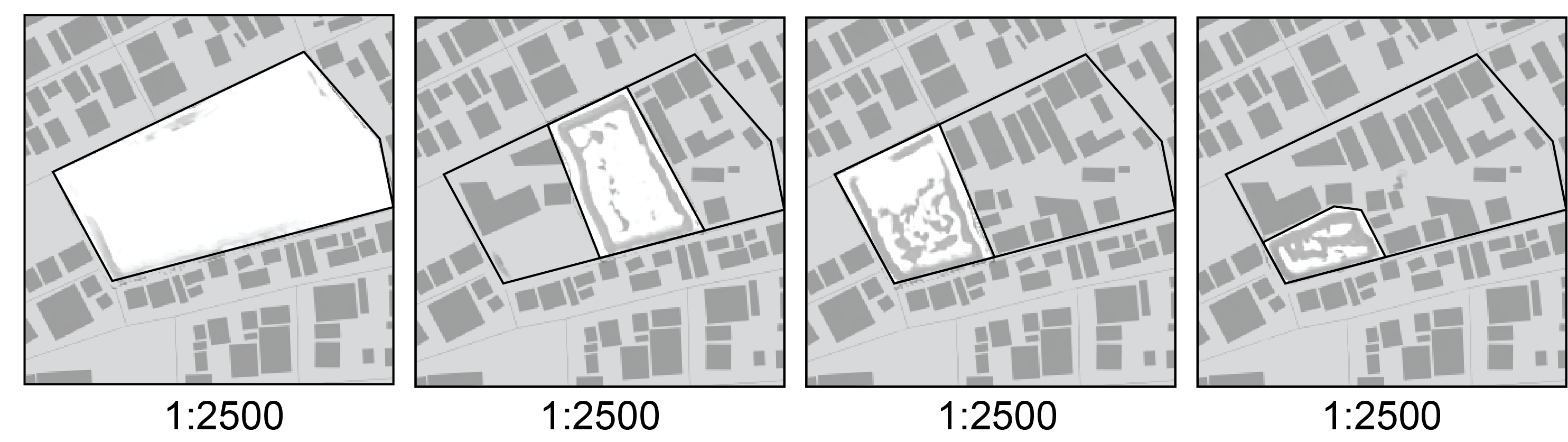}
    \caption{Illustration of the model's performance on the zone of interest with different proportions with respect to the image within the same Bengaluru urban context.}
    \label{fig:proportions}
    \Description{4 images with a diagrammatic representation of the city top view in scale 1:2500 with different proportion of the area defined as a block of interest. This shows that the model fails when the proportion of the block with respect to the image is large.}
    \centering
\end{figure}

Furthermore, the model was tested on a set of images from the same locations but with different scales to see whether the design capabilities depend on the scale and the proportions of the zone's size with respect to the surroundings. As it can be observed in Figure~\ref{fig:scales} performance of the model differs based on the scale. At the same time, it should be noticed that difference in performance is not related only to the scale but also to the proportion of the zone of interest: the block of Bengaluru (see Figure~\ref{fig:scales} second row, first column) at the scale of 1:2500 causes model's failure as the area of the block is too large with respect to the surroundings' area. On the scale of 1:5600, however, the performance is close to the one of Milan's sample at the scale of 1:2500. The areas of the respective blocks are similar. The same observation can be made for the last sample of Bengaluru and the second sample of Milan. Besides, the model was checked on an image on a large scale cropping the block to different proportions with respect to surroundings (Figure~\ref{fig:proportions}). The experiment demonstrates that the performance of the model depends on the proportion of the surroundings in an input image not on the urban scale and the fact that the design is conditioned by the context. This is the reason the importance of the presence of the context should be noted: firstly, the empty area is likely to be treated as the construction site and, secondly, there will be no surrounding blocks to condition the design of the current block of interest (see model's behavior in Figure~\ref{fig:milan_test} in samples 3, 6).
\newline However, there is another limitation related to the chosen training process and the generated images. The dimensions of the generated buildings remain the same independently from the scale of the input image (for instance, in Figure~\ref{fig:scales}, row 2): the width of the building volumes does not change with the modification of the scale of the surroundings. This behavior was expected, as all the images in the training set were produced on one scale, thus the distances and the units in the images are perceived by the model within this scale only.

\section{Style Translation}
Having obtained the results from the model trained on the city of Milan, it was decided to experiment further and see which features the model will learn from a completely different morphology. For this reason, several cities with a morphology that is distant from the morphology of Milan have been chosen:
\begin{itemize}
    \item Amsterdam
    \item Bengaluru
    \item Tallinn
    \item Turin
\end{itemize}
\begin{figure}[!h]
    \centering
    \includegraphics[width=0.9\columnwidth]{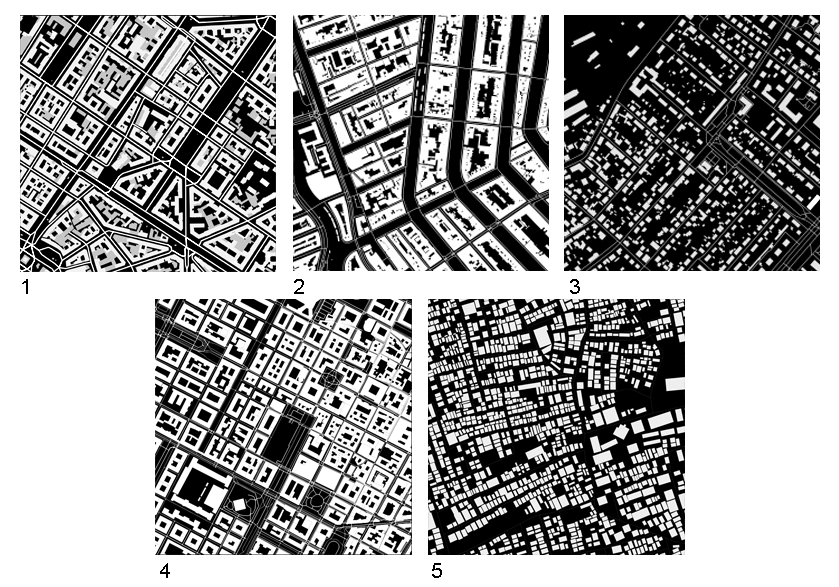}
    \caption{Illustration of the difference in urban morphology of the chosen cities. 1 - Milan, 2 - Amsterdam, 3 - Tallinn, 4 - Turin, 5 - Bengaluru}
    \Description{Five images showing the morphology of the following cities: Amsterdam, Bengaluru, Tallinn, Turin, Milan. The morphology is shown by drawing white building footprints from a part of each city on the black background.}
    \label{fig:morphology}
\end{figure}
The datasets for these cities are built with the use of scripting and QGIS software in the same way as the dataset of Milan urban blocks. The cities of Tallinn and Bengaluru had a condition of having more than 8 buildings in the block of interest due to the composition of their urban fabric.

As can be seen in Figure~\ref{fig:morphology} the cities vary greatly in urban density, buildings' types (isolated structures in Bengaluru and Tallinn; in connected structures and internal courtyard forming in Milan, Turin, and Amsterdam), relationships between the buildings and the block they belong to, the composition of the blocks. For instance, Milan, Turin, and Amsterdam have closed block structure often with an internal courtyard; at the same time the density of the blocks vary from city to city as well as the composition of the blocks: in Amsterdam typically a closed block is formed by many buildings while in Turin just by a few and the shape of a footprint of a block is more clear and geometrical. The differences in the urban fabric of the cities are helpful in the estimation of the learning results of the model and comparison of the results produced.

After the training and testing of these models on the corresponding cities each model was tested on the urban contexts from the other cities. This test gives an understanding of whether it is possible to translate an urban style from one city to another with the use of GAN, whether the model is able to adapt to completely different surroundings, and the coherence of the results with the urban blocks of training. The results were presented as a confusion matrix in Figure~\ref{fig:confusion_matrix}.
\newline The resulting models are published\cite{urbangen} and are available in open access under MIT License. The code for model testing is provided as well.

\section{Estimation}
The result of training for the case of urban block design should be evaluated both quantitatively and qualitatively. The qualitative approach is based on the use of a neural network predicting the class for the given images. Unfortunately, there was no possibility to have a survey based on the experts' opinion to obtain the qualitative estimation. This method serves as an alternative to experts' judgment on the blocks' design as it exploits the visual comprehension and appearance of an urban block in general that allows to attribute it to a particular city. This neural network classifier is trained independently of GAN and it represents a Convolutional Neural Network with a resnet50 backbone. This classifier is trained to distinguish among the urban blocks of all the cities the models were trained on.

The quantitative approach is based on the analysis of the block design from the urban and morphological perspective relative to the city it is situated in. The numerical parameters determining the morphology of an urban block are taken into account: relationships between the buildings within a block, relationships between the buildings and the block, average individual parameters of the buildings.

\subsection{Qualitative Estimation}

\begin{figure}[!h]
    \centering
    \includegraphics[width=0.9\columnwidth]{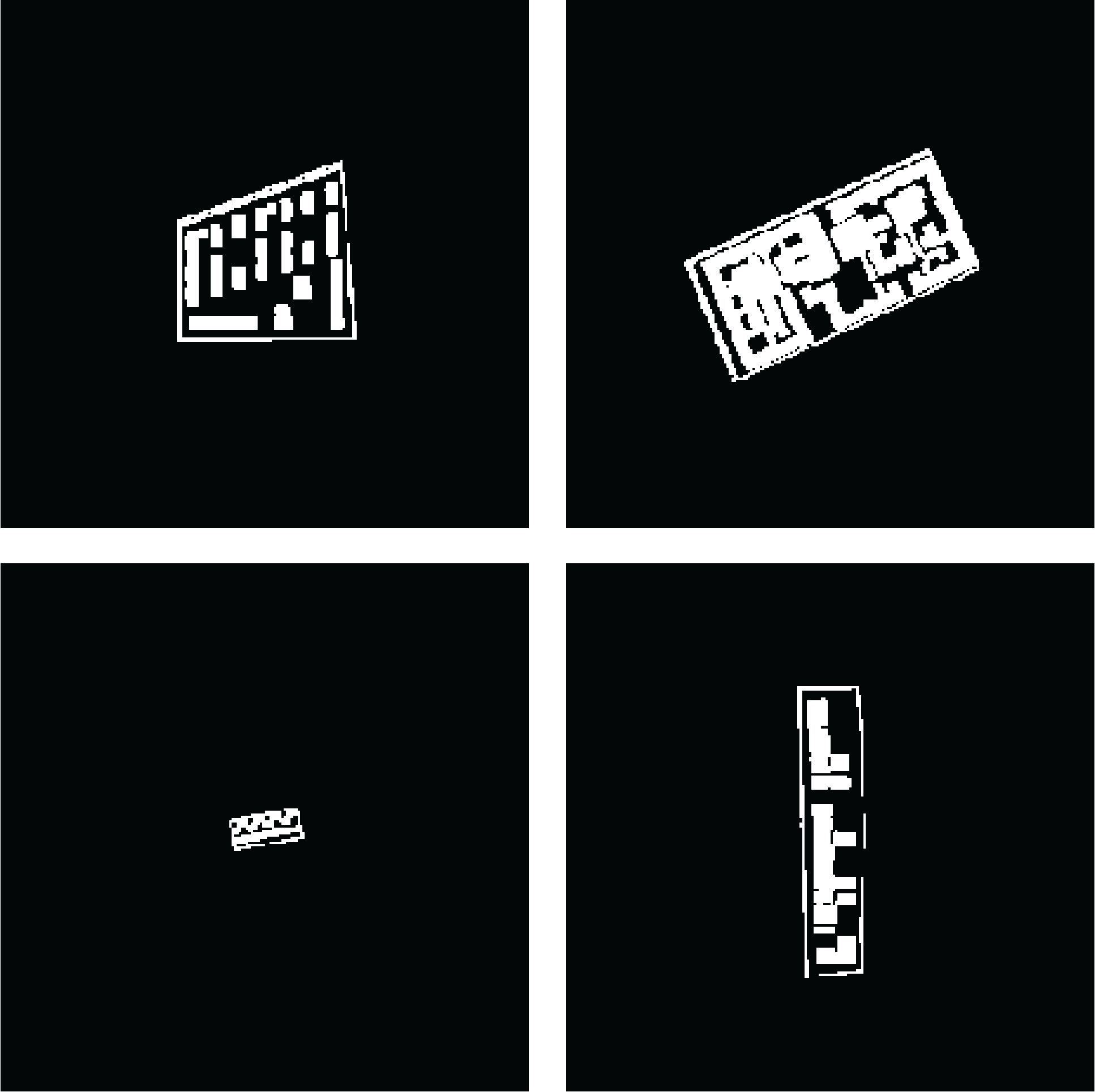}
    \caption{Samples from the dataset used for urban blocks classification.}
    \label{fig:classification_sample}
    \Description{Four samples from the dataset used for classification purposes. Each sample contains a white mask of an urban block on the black background.}
\end{figure}

A separate convolutional neural network with a resnet50 backbone \cite{DBLP:journals/corr/HeZRS15} was trained to classify between the images of the urban blocks of all the five cities. The training and validation sets were composed of the images from the real cities' structures \cite{github} which can be seen in Figure~\ref{fig:classification_sample} while the test images were the images generated by the GAN model. The results of the classification are reported in the Table ~\ref{tab:class_results}. The accuracy achieved on real samples is compared with the accuracy achieved on the generated samples.

\begin{figure*}[th!]
    \centering
    \includegraphics[width=1.0\textwidth]{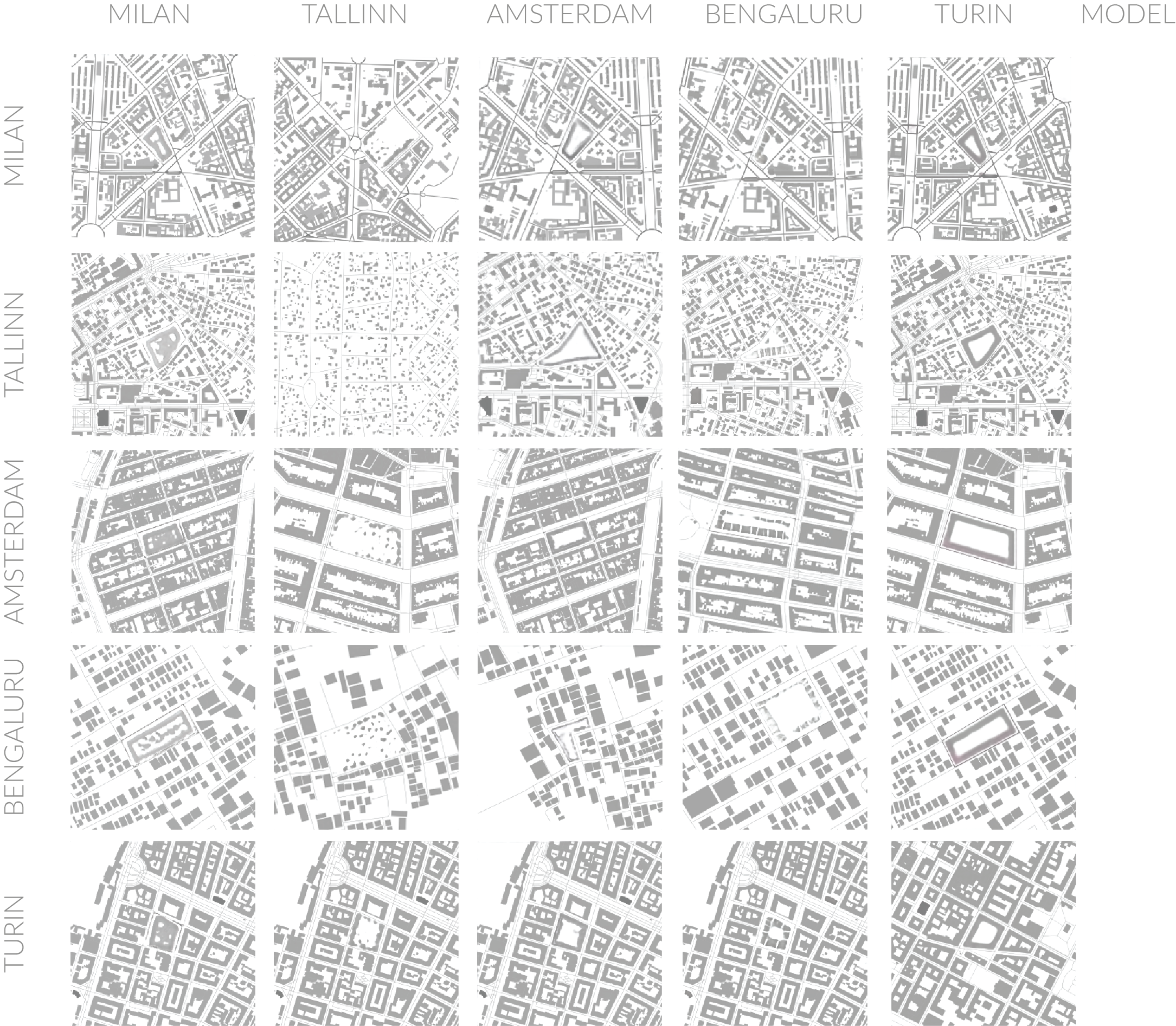}
    \centering
    \caption{Results obtained on the test set. Rows represent the existing urban context of the city, columns represent the cities the models were trained on (the expected morphology of the urban block). E.g. cell [1, 3], row 1 column 3 represents the model trained on the city of Amsterdam tested within Milan urban context: Amsterdam urban block design in Milan. The diagonal cells represent the models trained and tested on the same city.}
    \Description{A 5x5 matrix where every cell has an image of a city plan as a diagrammatic representation with different urban block designs produced by the generative models.}
    \label{fig:confusion_matrix}
    \centering
\end{figure*}

As in the case with the datasets for GAN, the classification dataset is divided into training and validation sets. Training images are the images the CNN learns from while validation images are used to see the results of the training process, these are the data the algorithm has not seen before. From the confusion matrix of results produced from the training set as seen in Table~\ref{tab:train_matrix} and validation set (Table~\ref{tab:valid_matrix}), it can be seen that urban blocks of some cities are confused more often with some specific cities. For example, Bengaluru and Turin do not have a large amount of misclassified images while the samples from Milan are often predicted to be from Tallin and vice versa.
\begin{table}[!h]
  \centering
  \caption{Classification Confusion Matrix from Training set results. Columns represent the predicted classes, rows represent the actual classes. E.g. cell [1, 3] row Amsterdam column Milan shows the number of Amsterdam sample images classified as Milan images.}
  
  \begin{tabular}{cccccc}
	\toprule
	 &
	\multicolumn{1}{p{0.1\columnwidth}}{\centering{Amster.}} &
    \multicolumn{1}{p{0.15\columnwidth}}{\centering{Bengaluru}} &
    \multicolumn{1}{p{0.1\columnwidth}}{\centering{Milan}} &
    \multicolumn{1}{p{0.1\columnwidth}}{\centering{Turin}} &
    \multicolumn{1}{p{0.1\columnwidth}}{\centering{Tallinn}} \\
    \midrule
    Amsterdam & 926 & 21 & 47 & 20 & 6 \\
    Bengaluru & 16 & 951 & 24 & 14 & 7 \\
    Milan & 55 & 22 & 866 & 14 & 38\\
    Tallinn & 54 & 26 & 32 & 484 & 10 \\
    Turin & 19 & 11 & 47 & 8 & 570\\
    \bottomrule
  \end{tabular}
  \label{tab:train_matrix}
 \end{table}
 
  \begin{table}[h]
  \centering
  \caption{Classification Confusion Matrix from Validation set results. Columns represent the predicted classes, rows represent the actual classes.}
  \begin{tabular}{cccccc}
	\toprule
	  &
	\multicolumn{1}{p{0.1\columnwidth}}{\centering{Amster.}} &
    \multicolumn{1}{p{0.15\columnwidth}}{\centering{Bengaluru}} &
    \multicolumn{1}{p{0.1\columnwidth}}{\centering{Milan}} &
    \multicolumn{1}{p{0.1\columnwidth}}{\centering{Turin}} &
    \multicolumn{1}{p{0.1\columnwidth}}{\centering{Tallinn}} \\
    \midrule
    Amsterdam & 146 & 3 & 0 & 17 & 13 \\
    Bengaluru & 1 & 167 & 0 & 12 & 3 \\
    Milan & 8 & 13 & 102 & 31 & 44 \\
    Tallinn & 19 & 2 & 4 & 82 & 1 \\
    Turin & 4 & 0 & 8 & 0 & 77 \\
    \bottomrule
  \end{tabular}
  \label{tab:valid_matrix}
\end{table}
 \begin{table}[h!]
  \centering
  \caption{Statistics from the classification model trained on the images of the urban blocks of Milan, Amsterdam, Bengaluru, Tallinn, Turin. The test accuracy is the confidence of prediction over the generated images from the Generator's model training set.}
  \begin{tabular}{ccccc}
    \toprule
    Classification &
    \multicolumn{1}{p{0.15\columnwidth}}{\centering{Training Accuracy}} &
    \multicolumn{1}{p{0.15\columnwidth}}{\centering{Validation Accuracy}} &
    \multicolumn{1}{p{0.15\columnwidth}}{\centering{Accuracy on Test Gen samples}} &
    \multicolumn{1}{p{0.15\columnwidth}}{\centering{Confidence On Test Gen Samples}} \\
    \midrule
    
    Amsterdam & 0.815 & 0.907 & 0.952 & 0.791 \\
    Bengaluru & 0.913 & 0.94 & 0.352 & 0.799 \\
    Milan & 0.515 & 0.87 & 0.619 & 0.877 \\
    Tallinn & 0.759 & 0.799 & 0.523 & 0.85 \\
    Turin & 0.865 & 0.87 & 0.5 & 0.795 \\
    Overall & 0.758 & 0.885 & 0.566 & 0.822 \\
    \bottomrule
  \end{tabular}
  \label{tab:class_results}
\end{table}

This classification model has been tested on the images generated by GAN. The test set consists of 50 images, 10 images per city. The results produced from Classifier over this set are reported in Table~\ref{tab:class_results}. As can be seen, the generated images are classified with more than 0.5 accuracy to the correct class apart from Bengaluru samples with a confidence of more than 0.75. Bengaluru model sometimes fails to produce a solution in Tallinn blocks due to the reason of low density.

\subsection{Quantitative Estimation}
Taking into consideration the city of Milan it can be noticed that from the morphological point of view its blocks could be divided into a few groups sharing some common features across the city based on the following characteristics: shape, dimension, number of parcels, and topography\cite{Biraghi:2018:ISBN:978-5-7638-4127-5}. While the first two are the conditions determined by the input image, the focus is on the latter two. The number of parcels and the topography of the blocks could be described numerically as parameters describing the relationships between the urban forms. These relationships have the same importance for the other cities as well. For this evaluation, the factors considered are the relationships between the buildings within a block (distance between the neighboring buildings), the relationships between a block and its buildings (building density, the proportion of built space in a block; the distance between the street and the closest building in a block) and the individual parameters of the buildings (average area of a building footprint in a block). We acknowledge that these parameters vary across one city, however, it is possible to deduce the most common characteristics shared by the majority of the blocks by taking the median value across the city. These most common characteristics represent the urban fabric, the feature that distinguishes one city from another~\ref{fig:morphology}. Moreover, in our case, the "inter-city" similarity is higher than the "intra-city" similarity as the cities have been chosen to have a 'contrasting' morphology. However, this research is limited by the number of comparison parameters for urban blocks; it would be a valuable continuation of the work to develop a complete set of comparison criteria.

\begin{table}
  \centering
  \caption{Quantitative estimation of the solutions produced by 5 trained GAN models: comparison of the average parameters by city and the parameters in the samples produced by the models.}
  \begin{tabular}{ccccc}
    \toprule
    City &
    \multicolumn{1}{p{0.11\columnwidth}}{\centering{City Dataset: Median Density}} &
    \multicolumn{1}{p{0.13\columnwidth}}{\centering{Generated Blocks: Median Density}} &
    \multicolumn{1}{p{0.15\columnwidth}}{\centering{City Dataset: Median Building Area, px}} &
    \multicolumn{1}{p{0.14\columnwidth}}{\centering{Generated Blocks: Median Building Area, px}} \\
    \midrule
    Amsterdam & 0.537 & 0.331 & 104 & 124 \\
    Bengaluru & 0.484 & 0.307 & 112 & 69 \\
    Milan & 0.489 & 0.521 & 258 & 215 \\
    Tallinn & 0.236 & 0.307 & 13 & 31 \\
    Turin & 0.553 & 0.594 & 283 & 389 \\
    \bottomrule
  \end{tabular}
  \label{tab:block_metrics}
\end{table}
\begin{table}
  \caption{Quantitative estimation of the solutions produced by 5 trained GAN models: comparison between the offset of the buildings from the street by city and the parameters in the samples produced by the models. Distance from the street is a minimum distance between the buildings in a block and the surrounding streets; the resulting value is a median value for the dataset samples and the samples produced by the models.}
  \centering
  \begin{tabular}{ccc}
    \toprule
    City &
    \multicolumn{1}{p{0.30\columnwidth}}{\centering{City Dataset: Median Distance Between the Buildings and the Street, px}}&
    \multicolumn{1}{p{0.35\columnwidth}}{\centering{Generated Blocks: Median Distance Between the Buildings and the Street, px}} \\
    \midrule
    Amsterdam & 1.14 & 1.414 \\
    Bengaluru & 1.01 & 3.4 \\
    Milan & 0.346 & 2.0 \\
    Tallinn & 1.85 & 2.4 \\
    Turin & 1.23 & 2.8 \\
    \bottomrule
  \end{tabular}
  \label{tab:street_distance}
\end{table}
\begin{table}
  \centering
  \caption{Quantitative estimation of the solutions produced by 5 trained GAN models: comparison between the median distance between the adjacent buildings by city and the parameters in the samples produced by the models. The distance between the adjacent buildings is an average of minimum distances between the buildings in a block; the resulting value is a median value.}
  \begin{tabular}{ccc}
    \toprule
    City &
    \multicolumn{1}{p{0.33\columnwidth}}{\centering{City Dataset: Median Distance Between the Adjacent Building}}&
    \multicolumn{1}{p{0.33\columnwidth}}{\centering{Generated Blocks: Median Distance Between the Adjacent Buildings}} \\
    \midrule
    Amsterdam & 2.6 & 2.5 \\
    Bengaluru  & 2.8 & 3.0 \\
    Milan & 1.5 & 2.2 \\
    Tallinn & 3.1 & 3.4 \\
    Turin & 0.5 & 0.0 \\
    \bottomrule
  \end{tabular}
  \label{tab:building_distance}
\end{table}
These parameters have been measured for all the blocks in each city dataset, then the median value has been calculated for each city. The median has been chosen over the mean value as it takes into consideration the presence of the outliers which are likely to be present in the datasets due to the urban variety in each city. The metrics have been calculated from the images with the use of Computer Vision\cite{github} and QGIS; the units of absolute values such as the median building area or the distance between adjacent buildings are pixels.

\begin{figure}[h!]
    \centering
    \includegraphics[width=0.95\columnwidth]{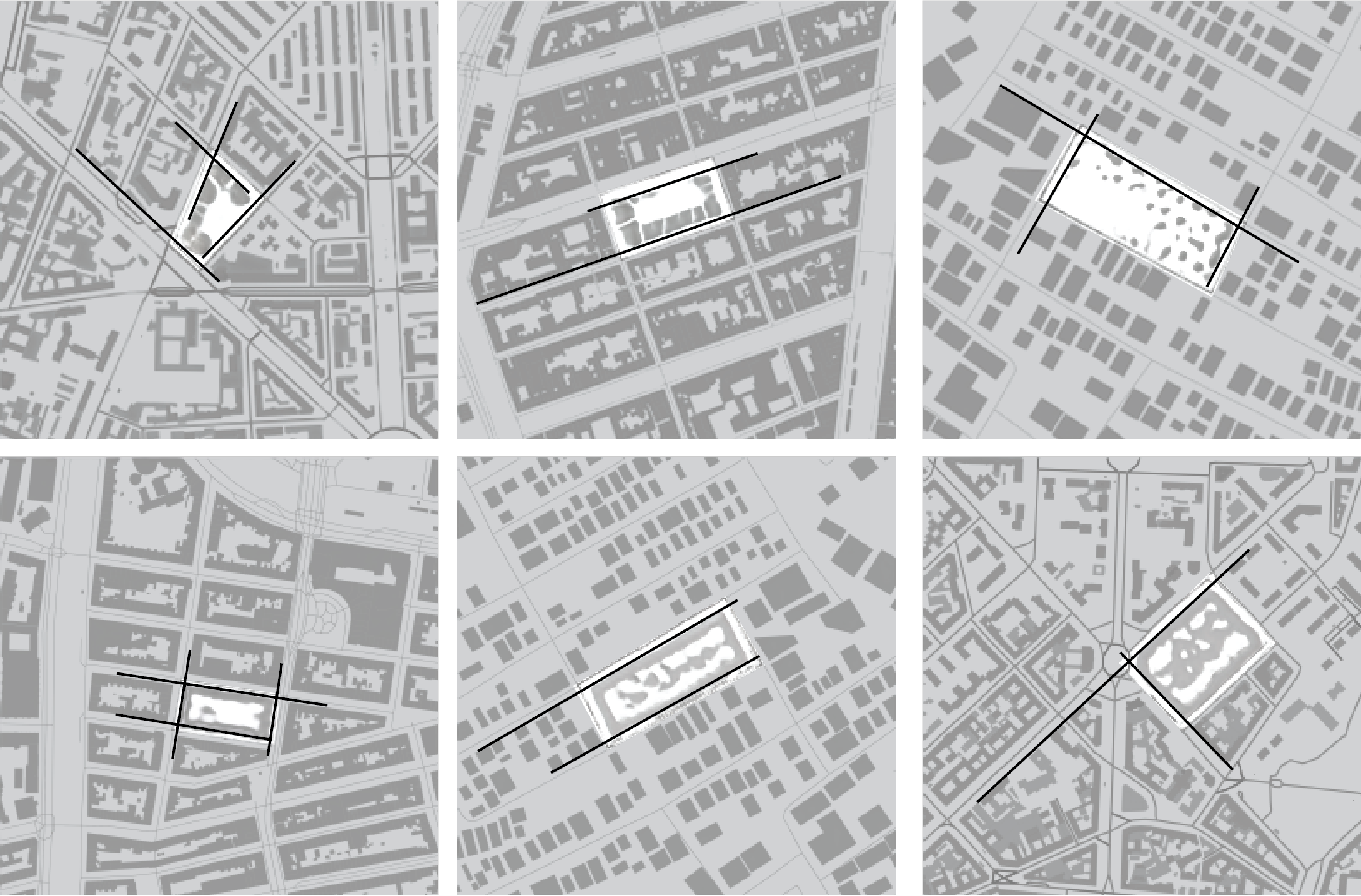}
    \caption{Samples from the test set. The block design is aligned with the surrounding blocks' street lines.}
    \Description{Six results from the test set with the annotations showing an alignment of the generated buildings to the existing buildings.}
    \label{fig:street}
\end{figure}

From Tables~\ref{tab:block_metrics},~\ref{tab:building_distance} it can be seen that urban blocks produced by the GAN in different urban contexts are coherent with the city of training: the parameters of building density, building area as in Table~\ref{tab:block_metrics} are close to the actual values, which confirms the observations made in the confusion matrix (Figure~\ref{fig:confusion_matrix}). At the same time, the distance from the street to the closest building in a block differs as in Table~\ref{tab:street_distance}: it is a constraint taken from the surroundings present in the input image (Figure~\ref{fig:street}).

\section{Contribution}
This paper aims to contribute to the field of generative design by:
\begin{itemize}
    \item introducing a new approach towards the context-based generative design of an urban block without a predefined set of constraints, by learning the visual parameters from the existing environment instead;
    \item introducing an application of an Image-to-Image translation GAN to the field of urban design and evaluating the produced design solutions from quantitative and qualitative points of view.
\end{itemize}
\section{Results}

The study shows that it is possible to achieve a generation of an urban block without the explicit definition of its parameters with the use of GAN. Not only does this method allow for learning the parameters from the existing urban context without their explicit definition, but it also produces results coherent with the urban surroundings of a block as in Figure~\ref{fig:confusion_matrix}. Moreover, this method produces a design solution with the characteristics of a generated urban block very close to the actual parameters of the blocks of a city of interest.

The experiments have shown that the model is applicable to diverse urban contexts with different urban morphology. Besides, urban parameters and morphology learned by a model can be translated to another city with a completely different context. The algorithm adapts the parameters learned previously to a new urban surrounding.

Some limitations of the algorithm include the size of an urban block within the image, the presence of the urban surroundings in the image, the scale of the buildings in the image. The size of a block is directly connected to the presence of the urban surroundings: the block design is shaped by the context, furthermore, the algorithm learns from the visible context. Thus it is important to have it present in a considerable portion of the image (determined experimentally). The scale of an urban block is limited due to the specifics of the current training preset: all the images from the training set were of the same scale, which means it is reasonable to test the algorithm with the images of the same scale. The preset is limited to a certain scale (which may vary from city to city based on the average block size) due to the necessity of having a sufficient part of the image filled with urban surroundings of a block of interest.

Besides, it should be noted that the design solutions produced do not have the same clarity as the vector images produced by the architects. However, urban shapes in the generated images are clearly visible and give a good understanding of the blocks' design. 

Another obvious limitation is the production of an urban block design that is similar to the existing blocks in the city. It is an approach to design that might be considered non-creative or lacking other important factors of urban design theory. We do not insist on the application of the suggested approach 'as-is'. This work highlights the ability of the algorithm to autonomously learn from the visual features, which could be further explored to take into account the other planning criteria and optimize the block's design from different points of view.

\section{Conclusion}
The experiment and evaluation of the GAN-based approach to the generative design of an urban block confirmed the possibility of the generation of a piece of urban fabric consistent with the surroundings and the desired urban style with the use of neural networks. Nonetheless, the images lack clarity of the vector images, the urban forms are compliant with the actual forms present in the city. What is more, these results have been achieved without image annotations, explicit definition of the parameters needed for each typology and shape of an urban block, typical characteristics of a city: all of these attributes are learnt by the neural network during the training process, from the experience having seen multiple urban surroundings.

In addition, the time and effort that are consumed for the block's design generation are much less compared to the human-generated design. It takes 2-4 hours to train one model after which it is able to generate the design solutions within 10 seconds.

This approach can not, obviously, substitute human participation in a city design process: the algorithm imitates the existing urban environment after having learned from it. It does not invent new space organization principles, does not rethink the comprehension of space by its users, and does not plan important landmarks of a city. However, considering that cities majorly consist of residential buildings and not landmarks, this approach could be used as a support tool for a planner or an architect by giving insights about the possible solutions and existing building composition of a city. Alternatively, it could be a design solution meant for typical blocks' design which would have a consistent integration into the existing surroundings. Moreover, as already mentioned, this approach could become a base for the consequent integration of the other urban design parameters into the model training which would result in a design that accustoms to multiple factors.

\begin{acks}
I would like to express my very great appreciation to Professor Luigi Mussio and Professor Federico Deambrosis who dedicated their time and effort to guiding and directing the work and encouraging me in the doubtful moments. My thanks also go to the professors of Architecture and Urban Design of Politecnico di Milano: Professor Lemes, Professor Caja, Professor Montedoro who had found time to give me their valuable suggestions from the urbanistic and architectural point of view as well as insights on the morphology of the cities. I would also like to thank Dr. Carlo Biraghi for the constructive advice.
\newline My gratitude is extended to Andrey Baran, Alberto Floris, Ram Kummamuru and Viktor Petukhov for the productive discussions and inspiration.
\end{acks}

\bibliographystyle{ACM-Reference-Format}
\bibliography{sample-base}


\begin{thebibliography}{22}


\ifx \showCODEN    \undefined \def \showCODEN     #1{\unskip}     \fi
\ifx \showDOI      \undefined \def \showDOI       #1{#1}\fi
\ifx \showISBNx    \undefined \def \showISBNx     #1{\unskip}     \fi
\ifx \showISBNxiii \undefined \def \showISBNxiii  #1{\unskip}     \fi
\ifx \showISSN     \undefined \def \showISSN      #1{\unskip}     \fi
\ifx \showLCCN     \undefined \def \showLCCN      #1{\unskip}     \fi
\ifx \shownote     \undefined \def \shownote      #1{#1}          \fi
\ifx \showarticletitle \undefined \def \showarticletitle #1{#1}   \fi
\ifx \showURL      \undefined \def \showURL       {\relax}        \fi
\providecommand\bibfield[2]{#2}
\providecommand\bibinfo[2]{#2}
\providecommand\natexlab[1]{#1}
\providecommand\showeprint[2][]{arXiv:#2}

\bibitem[\protect\citeauthoryear{??}{cc_}{[n.d.]}]%
        {cc_license}
 \bibinfo{year}{[n.d.]}\natexlab{}.
\newblock \bibinfo{title}{Creative Commons License. Attribution 2.5 Italy (CC
  BY 2.5 IT)}.
\newblock
  \bibinfo{howpublished}{\url{hhttps://creativecommons.org/licenses/by/2.5/it/deed.en}}.
\newblock


\bibitem[\protect\citeauthoryear{??}{pix}{[n.d.]a}]%
        {pixtorch}
 \bibinfo{year}{[n.d.]}\natexlab{a}.
\newblock \bibinfo{title}{Pix2Pix Implementation. Pytorch}.
\newblock
\newblock
\newblock
\shownote{\url{https://github.com/mrzhu-cool/pix2pix-pytorch}.}


\bibitem[\protect\citeauthoryear{??}{pix}{[n.d.]b}]%
        {pixtf}
 \bibinfo{year}{[n.d.]}\natexlab{b}.
\newblock \bibinfo{title}{Pix2Pix Implementation. Tensorflow}.
\newblock
\newblock
\newblock
\shownote{\url{https://github.com/affinelayer/pix2pix-tensorflow}.}


\bibitem[\protect\citeauthoryear{??}{rif}{[n.d.]}]%
        {riformaremilano}
 \bibinfo{year}{[n.d.]}\natexlab{}.
\newblock \bibinfo{title}{Riformare Milano}.
\newblock
\newblock
\newblock
\shownote{\url{http://www.riformaremilano.polimi.it/?page_id=1754}.}


\bibitem[\protect\citeauthoryear{Beirão, Mendes, Duarte, and Stouffs}{Beirão
  et~al\mbox{.}}{2010}]%
        {rule_based_design}
\bibfield{author}{\bibinfo{person}{José Beirão}, \bibinfo{person}{Gelly
  Mendes}, \bibinfo{person}{Jose Duarte}, {and} \bibinfo{person}{Rudi
  Stouffs}.} \bibinfo{year}{2010}\natexlab{}.
\newblock \showarticletitle{Implementing a Generative Urban Design Model:
  Grammar-based design patterns for urban design}.
\newblock \bibinfo{journal}{\emph{FUTURE CITIES [28th eCAADe Conference
  Proceedings / ISBN 978-0-9541183-9-6] ETH Zurich (Switzerland) 15-18
  September 2010, pp.265-274}}.
\newblock


\bibitem[\protect\citeauthoryear{Biraghi, Caja, and Zaroulas}{Biraghi
  et~al\mbox{.}}{2018}]%
        {Biraghi:2018:ISBN:978-5-7638-4127-5}
\bibfield{author}{\bibinfo{person}{Carlo~A. Biraghi}, \bibinfo{person}{Michele
  Caja}, {and} \bibinfo{person}{Sotirios Zaroulas}.}
  \bibinfo{year}{2018}\natexlab{}.
\newblock \showarticletitle{Urban Blocks and Architectural Typology in the
  Milanese Context}. In \bibinfo{booktitle}{\emph{Proc. XXV ISUF International
  Conference}}. \bibinfo{publisher}{Siberian Federal University Library},
  \bibinfo{pages}{859--868}.
\newblock


\bibitem[\protect\citeauthoryear{Boates}{Boates}{2020}]%
        {thismapdoesnotexist}
\bibfield{author}{\bibinfo{person}{Isaac Boates}.}
  \bibinfo{year}{2020}\natexlab{}.
\newblock \bibinfo{title}{This Map Does Not Exist project}.
\newblock
\newblock
\newblock
\shownote{\url{https://github.com/iboates/thismapdoesnotexist}.}


\bibitem[\protect\citeauthoryear{Chaillou}{Chaillou}{2020}]%
        {archigan}
\bibfield{author}{\bibinfo{person}{Stanislas Chaillou}.}
  \bibinfo{year}{2020}\natexlab{}.
\newblock \showarticletitle{ArchiGAN: Artificial Intelligence x Architecture}.
  In \bibinfo{booktitle}{\emph{Architectural Intelligence, Selected Papers from
  the 1st International Conference on Computational Design and Robotic
  Fabrication (CDRF 2019)}}. \bibinfo{pages}{117--127}.
\newblock
\showISBNx{978-981-15-6567-0}
\urldef\tempurl%
\url{https://doi.org/10.1007/978-981-15-6568-7_8}
\showDOI{\tempurl}


\bibitem[\protect\citeauthoryear{Chu, Li, Acuna, Kar, Shugrina, Wei, Liu,
  Torralba, and Fidler}{Chu et~al\mbox{.}}{2019}]%
        {ntg}
\bibfield{author}{\bibinfo{person}{Hang Chu}, \bibinfo{person}{Daiqing Li},
  \bibinfo{person}{David Acuna}, \bibinfo{person}{Amlan Kar},
  \bibinfo{person}{Maria Shugrina}, \bibinfo{person}{Xinkai Wei},
  \bibinfo{person}{Ming-Yu Liu}, \bibinfo{person}{Antonio Torralba}, {and}
  \bibinfo{person}{Sanja Fidler}.} \bibinfo{year}{2019}\natexlab{}.
\newblock \showarticletitle{Neural Turtle Graphics for Modeling City Road
  Layouts}.
\newblock  (\bibinfo{date}{10} \bibinfo{year}{2019}).
\newblock


\bibitem[\protect\citeauthoryear{{Comune di Milano}}{{Comune di
  Milano}}{2017}]%
        {Milano_geoportale}
\bibfield{author}{\bibinfo{person}{{Comune di Milano}}.}
  \bibinfo{year}{2017}\natexlab{}.
\newblock \bibinfo{title}{Milano Geoportale}.
\newblock
\newblock
\newblock
\shownote{\url{https://geoportale.comune.milano.it/sit/open-data/}.}


\bibitem[\protect\citeauthoryear{Deeplearning.ai}{Deeplearning.ai}{2020}]%
        {ganspecialization}
\bibfield{author}{\bibinfo{person}{Deeplearning.ai}.}
  \bibinfo{year}{2020}\natexlab{}.
\newblock \bibinfo{title}{GANs Specialization}.
\newblock
\newblock
\newblock
\shownote{\url{https://www.deeplearning.ai/generative-adversarial-networks-specialization/}.}


\bibitem[\protect\citeauthoryear{Fedorova}{Fedorova}{2020a}]%
        {github}
\bibfield{author}{\bibinfo{person}{Stanislava Fedorova}.}
  \bibinfo{year}{2020}\natexlab{a}.
\newblock \bibinfo{title}{Github Repository for urban datasets generation}.
\newblock
\newblock
\newblock
\shownote{\url{https://github.com/STASYA00/urban_datasets}.}


\bibitem[\protect\citeauthoryear{Fedorova}{Fedorova}{2020b}]%
        {urbangen}
\bibfield{author}{\bibinfo{person}{Stanislava Fedorova}.}
  \bibinfo{year}{2020}\natexlab{b}.
\newblock \bibinfo{title}{Github repository with published models}.
\newblock
\newblock
\newblock
\shownote{\url{https://github.com/STASYA00/UrbanGen}.}


\bibitem[\protect\citeauthoryear{Galanos}{Galanos}{2020}]%
        {daylightgan}
\bibfield{author}{\bibinfo{person}{Theodore Galanos}.}
  \bibinfo{year}{2020}\natexlab{}.
\newblock \bibinfo{title}{DaylightGAN}.
\newblock
\newblock
\newblock
\shownote{\url{https://github.com/TheodoreGalanos/DaylightGAN}.}


\bibitem[\protect\citeauthoryear{Goodfellow, Pouget-Abadie, Mirza, Xu,
  Warde-Farley, Ozair, Courville, and Bengio}{Goodfellow et~al\mbox{.}}{2014}]%
        {goodfellow2014generative}
\bibfield{author}{\bibinfo{person}{Ian Goodfellow}, \bibinfo{person}{Jean
  Pouget-Abadie}, \bibinfo{person}{Mehdi Mirza}, \bibinfo{person}{Bing Xu},
  \bibinfo{person}{David Warde-Farley}, \bibinfo{person}{Sherjil Ozair},
  \bibinfo{person}{Aaron Courville}, {and} \bibinfo{person}{Yoshua Bengio}.}
  \bibinfo{year}{2014}\natexlab{}.
\newblock \showarticletitle{Generative Adversarial Nets}. In
  \bibinfo{booktitle}{\emph{Advances in Neural Information Processing
  Systems}}, \bibfield{editor}{\bibinfo{person}{Z.~Ghahramani},
  \bibinfo{person}{M.~Welling}, \bibinfo{person}{C.~Cortes},
  \bibinfo{person}{N.~Lawrence}, {and} \bibinfo{person}{K.~Q. Weinberger}}
  (Eds.), Vol.~\bibinfo{volume}{27}. \bibinfo{publisher}{Curran Associates,
  Inc.}, \bibinfo{pages}{2672--2680}.
\newblock


\bibitem[\protect\citeauthoryear{He, Zhang, Ren, and Sun}{He
  et~al\mbox{.}}{2015}]%
        {DBLP:journals/corr/HeZRS15}
\bibfield{author}{\bibinfo{person}{Kaiming He}, \bibinfo{person}{Xiangyu
  Zhang}, \bibinfo{person}{Shaoqing Ren}, {and} \bibinfo{person}{Jian Sun}.}
  \bibinfo{year}{2015}\natexlab{}.
\newblock \showarticletitle{Deep Residual Learning for Image Recognition}.
\newblock \bibinfo{journal}{\emph{CoRR}}  \bibinfo{volume}{abs/1512.03385}
  (\bibinfo{year}{2015}).
\newblock
\showeprint[arxiv]{1512.03385}
\urldef\tempurl%
\url{http://arxiv.org/abs/1512.03385}
\showURL{%
\tempurl}


\bibitem[\protect\citeauthoryear{{Isola}, {Zhu}, {Zhou}, and {Efros}}{{Isola}
  et~al\mbox{.}}{2017}]%
        {isola2018imagetoimage}
\bibfield{author}{\bibinfo{person}{P. {Isola}}, \bibinfo{person}{J. {Zhu}},
  \bibinfo{person}{T. {Zhou}}, {and} \bibinfo{person}{A.~A. {Efros}}.}
  \bibinfo{year}{2017}\natexlab{}.
\newblock \showarticletitle{Image-to-Image Translation with Conditional
  Adversarial Networks}. In \bibinfo{booktitle}{\emph{2017 IEEE Conference on
  Computer Vision and Pattern Recognition (CVPR)}}.
  \bibinfo{pages}{5967--5976}.
\newblock
\urldef\tempurl%
\url{https://doi.org/10.1109/CVPR.2017.632}
\showDOI{\tempurl}


\bibitem[\protect\citeauthoryear{Nagy, Villaggi, and Benjamin}{Nagy
  et~al\mbox{.}}{2018}]%
        {inproceedings}
\bibfield{author}{\bibinfo{person}{Danil Nagy}, \bibinfo{person}{Lorenzo
  Villaggi}, {and} \bibinfo{person}{David Benjamin}.}
  \bibinfo{year}{2018}\natexlab{}.
\newblock \showarticletitle{Proc. SIMAUD 2018, Generative Urban Design:
  Integrating Financial and Energy Goals for Automated Neighborhood Layout}.
  \bibinfo{pages}{25}.
\newblock
\showISBNx{978-1-5108-6315-6}
\urldef\tempurl%
\url{https://doi.org/10.22360/simaud.2018.simaud.025}
\showDOI{\tempurl}


\bibitem[\protect\citeauthoryear{{OpenStreetMap contributors}}{{OpenStreetMap
  contributors}}{2017}]%
        {OpenStreetMap}
\bibfield{author}{\bibinfo{person}{{OpenStreetMap contributors}}.}
  \bibinfo{year}{2017}\natexlab{}.
\newblock \bibinfo{title}{{Data retrieved from https://www.openstreetmap.org
  }}.
\newblock \bibinfo{howpublished}{\url{ https://www.openstreetmap.org }}.
\newblock


\bibitem[\protect\citeauthoryear{{QGIS Development Team}}{{QGIS Development
  Team}}{2009}]%
        {QGIS_software}
\bibfield{author}{\bibinfo{person}{{QGIS Development Team}}.}
  \bibinfo{year}{2009}\natexlab{}.
\newblock \bibinfo{title}{QGIS Geographic Information System}.
\newblock
\newblock
\newblock
\shownote{\url{http://qgis.org}),.}


\bibitem[\protect\citeauthoryear{Wang, Fu, Wang, Huang, and Lu}{Wang
  et~al\mbox{.}}{2020}]%
        {landuseai}
\bibfield{author}{\bibinfo{person}{Dongjie Wang}, \bibinfo{person}{Yanjie Fu},
  \bibinfo{person}{Pengyang Wang}, \bibinfo{person}{Bo Huang}, {and}
  \bibinfo{person}{Chang-Tien Lu}.} \bibinfo{year}{2020}\natexlab{}.
\newblock \bibinfo{booktitle}{\emph{Reimagining City Configuration: Automated
  Urban Planning via Adversarial Learning}}.
\newblock \bibinfo{publisher}{Association for Computing Machinery},
  \bibinfo{address}{New York, NY, USA}, \bibinfo{pages}{497–506}.
\newblock
\showISBNx{9781450380195}
\urldef\tempurl%
\url{https://doi.org/10.1145/3397536.3422268}
\showURL{%
\tempurl}


\bibitem[\protect\citeauthoryear{Wang, Liu, Zhu, Tao, Kautz, and
  Catanzaro}{Wang et~al\mbox{.}}{2018}]%
        {nvidia}
\bibfield{author}{\bibinfo{person}{Ting-Chun Wang}, \bibinfo{person}{Ming-Yu
  Liu}, \bibinfo{person}{Jun-Yan Zhu}, \bibinfo{person}{Andrew Tao},
  \bibinfo{person}{Jan Kautz}, {and} \bibinfo{person}{Bryan Catanzaro}.}
  \bibinfo{year}{2018}\natexlab{}.
\newblock \bibinfo{title}{High-Resolution Image Synthesis and Semantic
  Manipulation with Conditional GANs}.
\newblock
\newblock
\newblock
\shownote{\url{https://tcwang0509.github.io/pix2pixHD/}.}


\end{thebibliography}










\end{document}